\pdfoutput=1

\documentclass[11pt]{article}

\usepackage{ACL2023}

\usepackage{times}
\usepackage{latexsym}
\usepackage{xcolor}
\usepackage{bm}

\usepackage[T1]{fontenc}

\usepackage[utf8]{inputenc}

\usepackage{microtype}

\usepackage{inconsolata}

\usepackage{tabularx}
\usepackage{booktabs}
\usepackage{amsmath}
\usepackage{graphicx}
\usepackage{multirow}
\usepackage{rotating}
\usepackage{float}
\usepackage{placeins}
\newcolumntype{L}[1]{>{\raggedright\let\newline\\\arraybackslash\hspace{0pt}}m{#1}}

\newcommand{\model}{{GlórIA}} 
\newcommand{\promptcolor}{teal}  
\newcommand{\gencolor}{darkgray}  

\title{\model: A Generative and Open Large Language Model for Portuguese\thanks{~~Pre-print - Accepted for publication at PROPOR 2024.}}

\author{Ricardo Lopes, João Magalhães, David Semedo \\
  NOVA LINCS, NOVA School of Science and Technology, Portugal \\
  \texttt{rv.lopes@campus.fct.unl.pt}\\
  \texttt{\{jmag, df.semedo\}@fct.unl.pt
 }
}

\begin{document}
\maketitle
\begin{abstract}
Significant strides have been made in natural language tasks, largely attributed to the emergence of powerful large language models (LLMs). These models, pre-trained on extensive and diverse corpora, have become increasingly capable of comprehending the intricacies of language. Despite the abundance of LLMs for many high-resource languages, the availability of such models remains limited for European Portuguese. We introduce \model, a robust European Portuguese decoder LLM. To pre-train \model, we assembled a comprehensive PT-PT text corpus comprising 35 billion tokens from various sources. We present our pre-training methodology, followed by an assessment of the model's effectiveness on multiple downstream tasks. Additionally, to evaluate our models' language modeling capabilities, we introduce CALAME-PT (Context-Aware LAnguage Modeling Evaluation for Portuguese), the first Portuguese zero-shot language-modeling benchmark.
Evaluation shows that \model~significantly outperforms existing open PT decoder models in language modeling and that it can generate sound, knowledge-rich, and coherent PT-PT text. The model also exhibits strong potential for various downstream tasks.
\footnote{For the source code, pre-trained models and data resources, refer to \url{https://github.com/rvlopes/GlorIA} and \url{https://huggingface.co/NOVA-vision-language}.}
\end{abstract}

\section{Introduction}
The emergence of robust large language models (LLMs) has led to a significant step forward across the whole natural language processing (NLP) field spectrum, with remarkable advances in a myriad of tasks, all of this with minimal supervision. Among the key ingredients to obtain such LLMs and enable effective modeling of language intricacies, we have 1) rich, highly diverse, and broad pre-training corpora accompanied by task-specific benchmarks to assess model capabilities in multiple down-stream tasks~\cite{GLUE,LAMBADA}; 2) high-capacity deep Transformer decoder architectures~\cite{attention_all_you_need,bloom}, and 3) state-of-the-art pre-training methodologies, to ensure stable convergence~\cite{biderman2023pythia,bloom}.

While such core language model learning ingredients have been thoroughly investigated and matured for English and other high-resource languages, the European Portuguese language is lagging behind.
In fact, there is a shortage of PT resources for pre-training and downstream task benchmarking, which is further aggravated when dialing down to European Portuguese (PT-PT). Additionally, it is critical to understand how well-established LLM learning methodologies, from data preparation and selection to training methodologies, generalize and ensure convergence on PT-PT corpora. 
Despite these limitations and challenges, there have been promising advances, with 
recent PT encoder models~\cite{albertinapt, BERTIMBAU} addressing many discriminative tasks with great success. However, many challenges remain open in Portuguese LLMs, in particular, in tasks that require language generation capabilities, in zero and few-shot settings, on a wide range of domains.

With these models and resource gaps in mind, we propose a new European Portuguese large decoder model, \textbf{\model}, trained on a diverse corpora comprising 35 billion tokens from a myriad of domains, including generic web content, news pieces, encyclopedic knowledge and dialog data.
Furthermore, to evaluate the language modeling capabilities of \model, we introduce CALAME-PT, a novel zero-shot PT benchmark for language modeling evaluation.
In our experiments, we show that \model~consistently and significantly outperforms existing PT open language models in language modeling.

\section{Related Work}
Generative LLMs have widely sparked the interest of the NLP community. All the way from GPT-2~\cite{gpt2} and GPT-3~\cite{GPT3}, new \textit{GPT-like} models have demonstrated impressive flexibility in addressing NLP tasks such as reading comprehension, question answering, among others, in zero and few-shot settings.
All these models adopt billion-scale parameter decoder-only Transformer architectures~\cite{attention_all_you_need,gpt2}, ranging from 1.3B up to 175B parameters. While each model uses its own pre-training corpora (some of which are not disclosed), the majority of the texts are in English.
It is particularly interesting the case of the LLaMA~\cite{LLAMA,Touvron2023Llama2O} family of models that are open and were trained with publicly available datasets, thus contributing to reproducibility.

\subsection{Moving towards PT LLMs}
With the goal of generalizing language knowledge, some initial multilingual models such as mBERT~\cite{bert}, mT5~\cite{mt5} and mGPT~\cite{mgpt} were contributed. Nevertheless, it has been shown that single-language LLMs outperform multilingual ones~\cite{camembert, finbert}.~\citet{BERTIMBAU} proposed BERTimbau, the first Brazilian Portuguese (PT-BR) encoder with that in mind. Moving towards larger encoder-decoder models,~\citet{ptt5} proposed the PTT5 model, based on the T5 architecture. It was then fine-tuned for paraphrasing tasks~\cite{ptt5paraphrasing}, and for Portuguese question-generation ~\cite{pt-questiongen}.

However, most of these are not exclusive to a specific variant of Portuguese, or lack generative capabilities, or are just fine-tuned to a specific downstream task. 
Focusing only on the PT-PT variant,~\citet{albertinapt} proposed Albertina, a 900M parameter DeBERTa encoder with both PT-PT and PT-BR versions, trained on different corpora due to language differences. The authors demonstrated that the PT-PT model outperformed its PT-BR counterpart on PT-PT tasks. The same authors also released the Gervásio-PTPT LLM, a 1B decoder available in HuggingFace\footnote{\url{https://huggingface.co/PORTULAN} - model name: \texttt{gervasio-ptpt-base}.}.
Recently, Sabiá~\cite{sabia}, a 65B PT-BR LLM based on LLaMA was proposed, showing promising results on PT-BR few-show settings. 

\subsection{Large-Scale PT Text Data}
Previously mentioned models leveraged large-scale unlabeled text data. Major efforts have been made to produce massive collections of text for heavily researched languages like English. For Portuguese, there have been some promising advances. BERTimbau used brWac~\cite{brwac}, a 2.7B token dataset obtained from crawling PT-BR websites, while Albertina used a PT-PT filtered version of OSCAR, together with PT-PT transcripts datasets from the Portuguese and the EU Parliaments~\cite{dcep,europarl}. Sabiá uses the Portuguese subset of ClueWeb22~\cite{CLUEWEB22}.
Leveraging filtered massive web crawls such as ClueWeb22~\cite{CLUEWEB22} and OSCAR~\cite{OSCAR}, the Portuguese web-archive~\cite{ArquivoPT} (Arquivo.pt), encyclopedic and dialog data, we assemble and contribute with a large and highly-diverse pre-training PT-PT corpus.

\begin{table*}[!htb]
\centering
\caption{Collected datasets and post-processing statistics.}
\begin{tabular}{llrr}
\toprule
\textbf{Dataset}            & \textbf{Domain}                          & \textbf{Documents} & \textbf{Tokens} \\ 
\midrule
\textbf{ClueWeb22 PTPT Subset}       & Web Crawl                                & 29M                      & 31.6B                      \\ 
\textbf{OSCAR PTPT}         & Web Crawl                                & 1.5M                     & 1.8B                             \\ 
\textbf{ArquivoPT}          & News and periodicals                     & 1.5M                     & 0.8B                             \\ 
\textbf{OpenSubtitles PTPT} & Subtitles from movies                    & 1.2M                     & 1.0B                      \\ 
\textbf{PTWiki}             & Encyclopedia                             & 0.8M                     & 0.2B                     \\
\textbf{EuroParl PTPT}      & European Parliament Dialogs & 1.3M                     & 0.05B                          \\ \midrule
 & \textbf{Total} & 35.3M & 35.5B \\
\bottomrule
\end{tabular}%
\label{tab:datasetspostprocess}
\end{table*}

\section{Preparing a new Large PT-PT Corpus}

As evidenced by previous work, a large and diverse collection of texts, spanning over multiple domains, allows the model to better understand the language (and its intrincacies), thus improving the quality of the generated text~\cite{gpt2,LLAMA,GPT3}.
Given that the availability of European Portuguese texts at scale is not on par with English, our first objective is to further advance the diversification and availability of PT-PT resources, by gathering a large and rich collection of datasets. 

\subsection{PT Language Sources}
To gather high-quality, large-scale, PT language resources, we resorted to multiple PT-PT text sources, summarized in Table~\ref{tab:datasetspostprocess}. \textbf{OSCAR-2201~\cite{OSCAR} and ClueWeb-L 22 ~\cite{CLUEWEB22}} are web crawls -- they both give us text from blogs, forums, among other websites. The \textbf{PTWiki}\footnote{\url{https://dumps.wikimedia.org/}} provides our model with well-written and reviewed encyclopedic knowledge, in neutral and revised Portuguese text. \textbf{Europarl}\footnote{\url{https://www.statmt.org/europarl/}}~\cite{europarl} provides transcripts from diverse sessions that occurred in the European Parliament (such as colloquial conversations between Eurodeputies). \textbf{OpenSubtitles}~\cite{opensubtitles2016} is comprised of essentially small and short movie conversations and narrations. 
Finally, our \textbf{Arquivo.pt subset} is a collection of scrapped text from periodicals and news websites archived by Arquivo.pt~\cite{ArquivoPT}, providing the model with high-quality reviewed news texts.

\subsection{Data Processing}
Once the individual datasets were gathered, they were filtered and processed. PT-PT documents were filtered using metadata when available (documents whose URL contains ".pt" in its domain), removing documents with low word count (<=15), fixing \textit{mojibakes} and other encoding errors, removing remnant HTML tags, and removing exact duplicates through hashing. 
To avoid having the model learn "first-person" toxicity biases and insults, an extra processing step was applied to OpenSubtitles to discard samples based on the existence of profanity words, where a manually produced list of Portuguese bad words was used to explicitly filter out samples that contained them. 
We believe that this had to be done specifically for OpenSubtitles due to its dialog nature as we wanted to avoid having the model learn "first-person" toxicity biases.

After processing, our pre-training corpus reached a total of 35.3M documents and 35.5B tokens -- Table \ref{tab:datasetspostprocess} shows the detailed statistics.

\section{The \model~Model }
\model~is a decoder-based LLM with an architecture similar to GPT-3's~\cite{GPT3}, competing with it in linguistic, physical, and scientific reasoning tasks. Specifically, it adopts the GPTNeo~\cite{GPTNEO}'s 1.3B and 2.7B architectures, following the HuggingFace's implementation of the model. Being a decoder, \model~uses a Causal LM pre-training objective, using cross-entropy as its loss.
Table \ref{table:ptptmodelsizecomp} shows the architecture configuration for \model's both versions.
GPTNeo also employs local attention~\cite{LocalAttention}, which replaces standard self-attention and combines a dilating sliding window strategy with pre-selected global attention on some input locations, making the self-attention scale linearly, and linear attention~\cite{LinearAttention}, which optimizes the dot-products by providing linear memory and processing complexities while maintaining representational capability.

\begin{table}[t]
\caption{\model~architecture configurations. $l$ denotes the number of layers, \#AH the number of attention heads, and $h$ denotes the model hidden layer size. }
\centering
\begin{tabular}{ccccc} 
\toprule
\textbf{Model} & \textbf{\#Params.} & \textbf{\bm{$l$}} & \textbf{\#AH} & \textbf{\bm{$h$}} \\
\midrule
\textbf{\model~1.3B} & 1.3B & 24 & 16 & 2048 \\
\textbf{\model~2.7B} & 2.7B & 32 & 20 & 2560 \\
\toprule
\end{tabular}
\label{table:ptptmodelsizecomp}
\end{table}

\begin{figure*}[!t]
 \centering
   \includegraphics[width=0.75\textwidth]{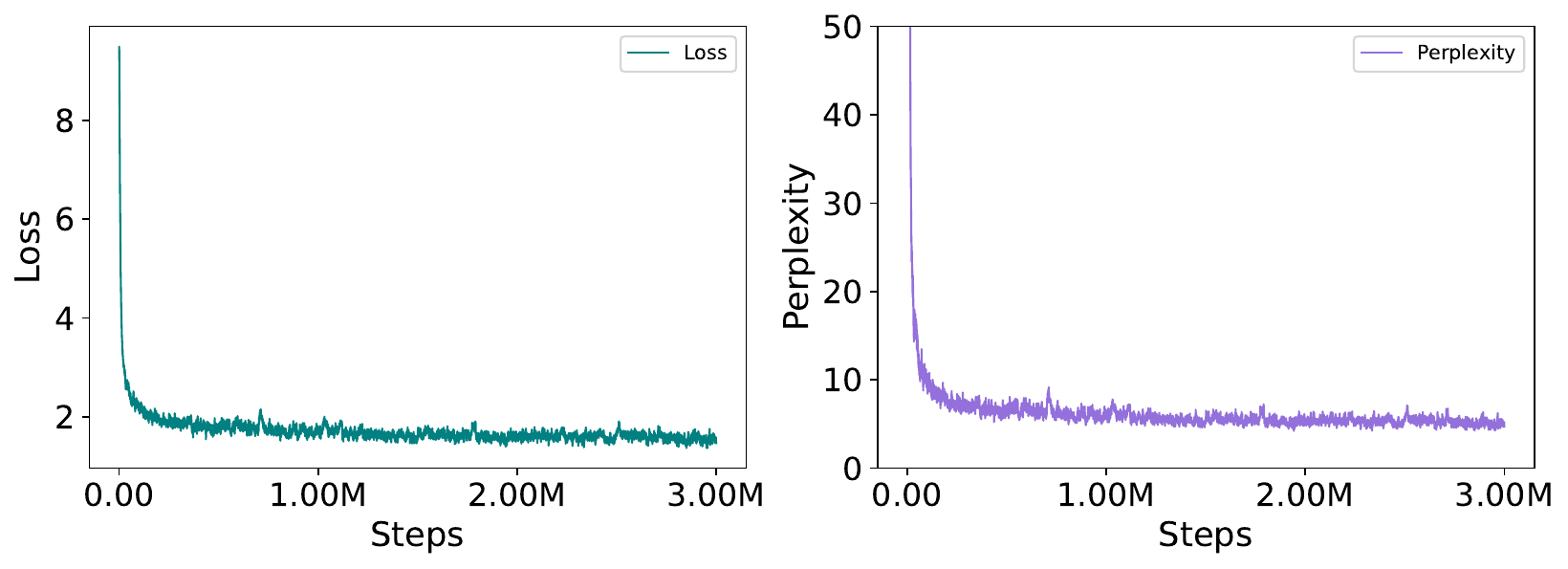}
  \caption{\model~1.3B pre-training loss and perplexity.}
  \label{fig:lossppl}
\end{figure*}

\subsection{Pre-training details}
To pre-train \model, a total batch size of 512 was used (128 p/ GPU), with 16 gradient accumulation steps.
We prepared a GPT-2-like BPE tokenizer, with a vocabulary size of 50257 tokens. Training was performed with BF-16 mixed-precision and a weight decay of 0.01. For the 1.3B version, \model~was trained for a total of 3M steps, on 4x NVIDIA A100s 40GB, for a total of 21 days (7 days p/ 1M steps), while, for the 2.7B, due to hardware resource constraints, we trained it only for 1M steps on 7x NVIDIA A100s (10 days).
A cosine annealing scheduler was used for both models, with hard restarts every 500k steps and 10k warmup steps. Periodic evaluations and data shuffling were conducted every 1 million steps.

\subsection{Data Sampling Strategy}
In order to take advantage of the diversity of our data, we implemented a sampling strategy similar to LLAMA's~\cite{LLAMA} where we attribute specific probabilities to each dataset, so that we can control which and how much data the model sees. In sum, a batch is prepared by sampling documents from every dataset according to pre-assigned sampling probabilities. Table \ref{tab:seendocs} presents the total data seen during the 1.3B model pre-training as well as the sampling probabilities $p(i)$. Higher probability was given to ClueWeb since it constitutes the bulk of our data. Thus, we decided to spread the remaining datasets with balanced percentages, akin to LLAMA's distribution. The same weights were used for the 2.7B version.
\begin{table}[t]
\centering
\caption{Documents seen per each dataset during \model~1.3B's pre-training - a total of 96M documents. \textbf{Seen Docs.} denotes the number of documents seen in training, \textbf{\#E} denotes the number of epochs of the corresponding subset, and $\bm{P(i)}$ denotes the probability of sampling a document $i$ from that subset.}
\begin{tabular}{lccc}
\toprule
\textbf{Dataset} &  \textbf{Seen Docs.} & \textbf{\#E} & $\bm{P(i)}$ \\ 
\midrule
\textbf{ClueWeb PTPT}  & 59.870M & 2.06 & 0.62 \\
\textbf{PTWiki} & 9.516M & 11.60 & 0.10 \\ 
\textbf{OSCAR PTPT} & 7.610M & 4.88 & 0.08 \\ 
\textbf{ArquivoPT}& 7.598M & 5.07 & 0.08 \\ 
\textbf{OpenSub. PTPT}& 5.707M & 4.41 & 0.06 \\ 
\textbf{EuroParl PTPT}  & 5.700M & 4.08 & 0.06 \\ \bottomrule
\end{tabular}%
\label{tab:seendocs}
\end{table}

\subsection{Training Convergence}

Figure~\ref{fig:lossppl} depicts the loss (at the left) and perplexity (at the right) evolution during pre-training, for the \model 1.3B variant. We start by observing a rapid loss decrease in the first 1 million steps. Then, a slower but steady decrease can be observed, until the end of the training.

\begin{table*}[!ht]
\caption{Examples of CALAME-PT's samples. We present the \textit{prompts} and the \textcolor{\promptcolor}{target words} the models should predict given the context, and if they're generated or handwritten.}
\centering
\begin{tabularx}{\linewidth}{X|X}
\toprule
\textbf{Handwritten (H):} \textit{Um gato andava atrás do rato mas não o conseguia apanhar. Para todo o lado o rato fugia e fugia e o gato não o conseguia apanhar. Até que o gato se conseguiu adiantar e finalmente comeu o}  \textcolor{\promptcolor}{rato}
& 
\textbf{Handwritten (H):} \textit{A tragédia atingiu a família quando ele caiu no chão e não havia ninguém no local com formação em primeiros}  \textcolor{\promptcolor}{socorros}
\\ \midrule
\textbf{Generated+Reviewed (A):} N\textit{o contexto apresentado, várias organizações do trabalho, como sindicatos e associações sindicais, estão envolvidas em negociações e revisões contratuais com várias empresas. Essas interações destacam a importância das negociações coletivas para garantir condições justas de trabalho. As organizações do trabalho trabalham em conjunto para representar os interesses dos}  \textcolor{\promptcolor}{trabalhadores} & 
\textbf{Generated+Reviewed (A):} \textit{Depois de um período de controvérsia, uma empresa decidiu suspender a partilha de dados de utilizadores para fins publicitários. A decisão foi tomada após protestos em diferentes países. A suspensão é temporária e a empresa está a trabalhar com as autoridades para retomar a partilha de dados. Esta situação levanta questões sobre a segurança e privacidade dos}  \textcolor{\promptcolor}{utilizadores} 
\\ \bottomrule
\end{tabularx}%
\label{tab:calamesamples}
\end{table*}

\section{Evaluation of PT Language Generation}
We introduce the first zero-shot Portuguese language modeling benchmark, CALAME-PT (Context-Aware LAnguage Modeling Evaluation for Portuguese). Inspired by the widely used LAMBADA~\cite{LAMBADA} benchmark, the task consists of guessing the final word given the context that comes before it. It comprises \textbf{a total of 2076 texts and respective last words}, covering a wide variety of domains and contexts, whose context should be enough to guess the word. The topic diversity and the zero-shot setting directly requires models to leverage their inner knowledge to correctly solve the task. The target word can either be present or not in the context, which should be enough to predict it.

\begin{figure}[t]
  \centering
    \includegraphics[width=0.5\textwidth]{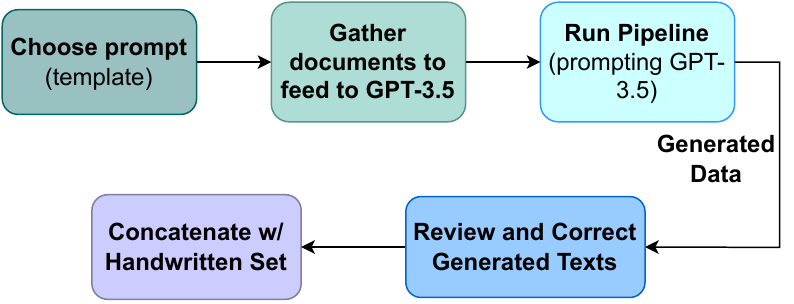}
  \caption{Overview of the CALAME-PT's generated set creation process.}
  \label{fig:calameprocess}
\end{figure}

\subsection{Building CALAME-PT}
When creating the CALAME-PT benchmark, an hybrid approach is used to strike a balance between scale and diversity (w.r.t. to different domains and difficulty).  
As such, we produced two sets of samples: one with fully handwritten samples (\textbf{H}) and one with automatic generation+human review samples (\textbf{A}).
For the handwritten set, a total of 406 samples were handwritten by 4 annotators, where it was sought to cover a broad set of domains.

For the automatic generation+human review, a pipeline was built to generate new texts grounded on a set of randomly sampled documents from a small subset of documents from ArquivoPT, PTWiki, and OSCAR were chosen, purposely left out from the training set. 
This was accomplished by prompting GPT-3.5 - the prompt is shown in Table~\ref{tab:chosenprompt} - and generating a total of 2.5k samples. This process cost $\approx 7$ euros. Then, these samples were human-reviewed to remove low-quality samples, anonymize samples, fix minor mistakes, and address ambiguity by performing small rewrites. 
In the end, we were left with 1670 generated samples. The handwritten and automatically generated+human reviewed sets were combined (\textbf{ALL}) to create the final version, resulting in a total of 2076 samples.

\begin{table}[t]
\caption{The chosen prompt that was fed to GPT3.5 to generate a new, smaller text based on our documents.}
\centering
\begin{tabular}{|L{7,3cm}|}
\hline
Dado o seguinte contexto: \\ 
< \textcolor{\promptcolor}{DOC HERE} > \\ 
Escreve um pequeno texto inspirado pelo contexto com poucas frases. Não deves mencionar nomes de pessoas ou países, eventos, marcas, e datas (dias, anos e horas). \\ \hline
\end{tabular}%
\label{tab:chosenprompt}
\end{table}

\subsection{Caveats of LLM-based Sample Creation}
When preparing CALAME-PT, the first difficulty was ensuring anonymization and removal of encyclopedic knowledge-dependent contexts, in order to make each samples' context self-contained. We asked GPT-3.5 to perform these steps but sometimes it would fail. The second issue was GPT-3.5's struggle to generate accurate European Portuguese text. While the generated text was generally correct, it had a tendency to shift to PT-BR, which led to the presence of sporadic PT-BR linguistic traits in some of the samples.
Another note is the on ambiguous contexts, which can  needlessly harm a model's performance (models may generate a word that makes sense but does not  exactly correspond to the target word). Thus, we aimed toward a sensible balance between the ambiguity and predictability present in the samples.

\subsection{Evaluation Protocol}
We compared our 1.3B (1M to 3M steps' checkpoints) variants of \model~to two decoder-based models: Gervásio-PTPT and mGPT~\cite{mgpt}. Gervásio-PTPT is based on the Pythia 1B model~\cite{biderman2023pythia}, and mGPT is a 1.3B  multilingual variant resembling the GPT-3 architecture.
We chose greedy and beam search + top-k decoding strategies for evaluation, with  4 beams, $k=50$, with a temperature of 1.0 and a token repetition penalty of 2. Due to its non-deterministic nature, we report the average of 3 runs.

The models were evaluated on the entire CALAME-PT dataset, in a zero-shot setting, followed by a separate evaluation of the handwritten (\textbf{H}) and automatically generated + human reviewed (\textbf{A}) sets. In practice, we have the models generate up to 5 new tokens, and we only consider the first full generated word. We then compare it against the ground-truth target last word, by ignoring casing and accents. 

\begin{table}[t]
\centering
\caption{CALAME-PT benchmark results (\textbf{Exact-Match}) comparison using the \textbf{greedy decoding strategy}.}
\resizebox{0.5\textwidth}{!}{%
\begin{tabular}{lccc}
\toprule
\textbf{Models} & \textbf{ALL} & \textbf{A} & \textbf{H} \\ 
\toprule
Gervásio-PTPT & 19.03 & 19.88 & 15.52 \\  
mGPT & 29.47 & 31.55 & 20.93 \\ \midrule
\model~1.3B (1M Chk) & 35.07 & 37.36 & 25.62 \\ 
\model~1.3B (2M Chk) & 35.93 & 38.14 & 26.84 \\ 
\textbf{\model~1.3B (3M Chk)} & \textbf{36.61} & \textbf{38.86} & \textbf{27.34} \\ 
\bottomrule
\end{tabular}%
}
\label{tab:calamegreedyresults}
\end{table}

\begin{table}[!t]
\centering
\caption{CALAME-PT benchmark results (\textbf{Exact-Match}) comparison using the \textbf{beam search with top-k sampling strategy}. Each score is the average of 3 runs.}
\begin{tabular}{lccc}
\toprule
\textbf{Models} & \textbf{ALL} & \textbf{A} & \textbf{H} \\ \toprule
Gervásio-PTPT & 44.01 & 45.97 & 34.90 \\ 
mGPT & 47.14 & 50.03 & 35.87 \\ \midrule
\model~1.3B (1M Chk) & 50.99 & 53.21 & 38.75 \\ 
\model~1.3B (2M Chk) & 51.80 & 53.69 & 41.95 \\ 
\textbf{\model~1.3B (3M Chk)} & \textbf{52.79} & \textbf{55.39} & \textbf{42.61} \\  
\bottomrule
\end{tabular}%
\label{tab:calamesamplingresults}
\end{table}

\subsection{CALAME-PT Results Discussion}

\begin{figure}[t]
  \centering
    \includegraphics[width=0.5\textwidth]{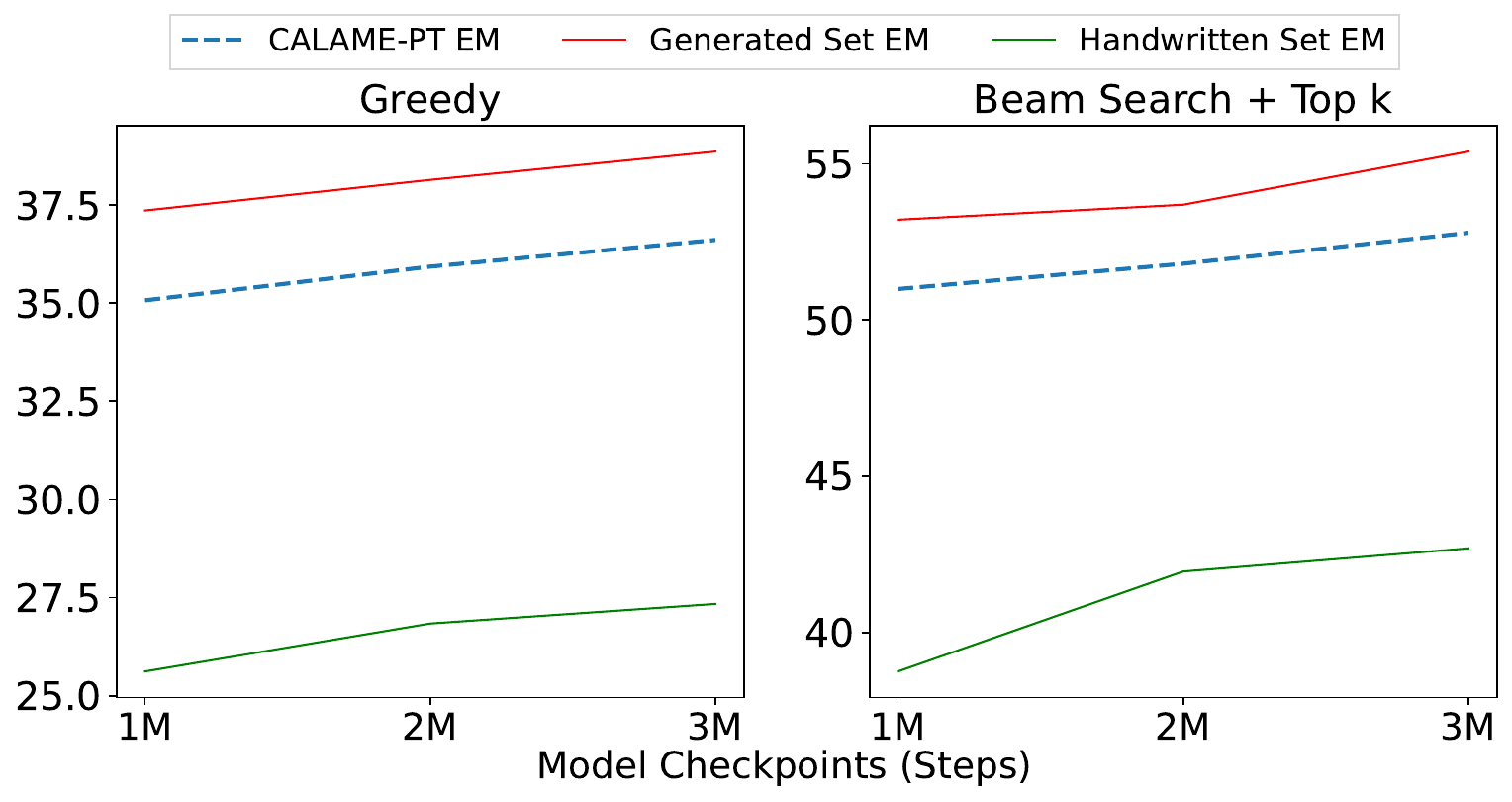}
  \caption{Evolution of \model~1.3B performance on CALAME-PT. Evaluated at 3 distinct checkpoints (1M, 2M, and 3M steps) for both decoding strategies. \textbf{EM} denotes Exact-Match.}
  \label{fig:calameevolution}
\end{figure}

\noindent\textbf{Overall Results.} We present the results for CALAME-PT for the greedy and beam search+top-k strategies in, respectively, Tables \ref{tab:calamegreedyresults} and \ref{tab:calamesamplingresults}. The first conclusion is that the beam-search + top-k sampling is significantly better for text generation, matching our initial qualitative observations. The second is that both versions of \model~outperform Gervásio-PTPT and mGPT by a relevant margin, in all settings. 
It can also be seen that training longer leads to a consistent performance improvement, with an observed $\approx 4\%$ relative improvement between the 1M and 3M checkpoints.
This is also supported by Figure \ref{fig:calameevolution}, which evidences the consistent performance evolution throughout training checkpoints.

\vspace{5pt}
\noindent\textbf{Results per Subset (H vs. A).} Regarding the results on each subset - handwritten (H) vs. automatically generated+human reviewed (A), it is interesting to see that samples from the H set are more challenging. In particular, we observe a $10\%$ performance drop in \model 1.3B (3M Chk), with both decoding strategies, compared to the A set. We posit that there is an inherent bias to GPT-3.5 generated samples, that leads to more predictable target words.

\begin{table}[!t]
\centering
\caption{CALAME-PT's generated set results (exact-match as percentage) discriminated by the source dataset used to create the samples (using beam search). \textit{PW} - PTWiki. \textit{Arq} - ArquivoPT. \textit{Osc} - OscarPTPT.}
\begin{tabular}{lccc}
\toprule
\textbf{Models} & \textbf{\textit{PW}} & \textbf{\textit{Arq}} & \textbf{\textit{Osc}} \\ \toprule
Gervásio-PTPT & 46.15 & 45.42 & 45.80 \\ 
mGPT & 49.69 & 50.08 & 50.57 \\ \midrule 
\model~1.3B (3M Chk) & 53.84 & 55.76 & 56.20 \\ 
\model~2.7B (1M Chk) & 54.61 & 54.06 & 55.89 \\
\bottomrule
\end{tabular}%
\label{tab:generatedsourceresults}
\end{table}

\begin{table}[t]
\centering
\caption{Comparison between \model~1.3B and ~\model~2.7B (EM), after 1M training steps, using beam search with top-k sampling. 
Each score is the average of 3 evaluations.
}
\begin{tabular}{lccc}
\toprule
\textbf{Models} & \textbf{ALL} & \textbf{A} & \textbf{H} \\ \toprule
\model~1.3B (1M Chk) & 50.99 & 53.21 & 38.75 \\ 
\model~2.7B (1M Chk) & 52.20 & 54.57 & 40.40 \\ \bottomrule
\end{tabular}%
\label{tab:calame_1b_2b}
\end{table}

\vspace{5pt}
\noindent\textbf{Results Per Source on the Automatically Generated set (A).} We recall that the automatically generated + human reviewed subset (A) was created by sampling documents from three different sources (ArquivoPT, PTWiki, OSCAR PT).
To understand the models' performance per source, we present in Table~\ref{tab:generatedsourceresults} the results, by discriminating by the samples' dataset source. 
The main observation is that performance is quite balanced over the three distinct sources, over all the compared models. We observe that for samples grounded in OSCAR PT, performance is consistently (but marginally) higher. For \model 1.3B and mGPT, samples grounded on PTWiki are the most challenging.

\subsection{Comparing 1.3B and 2.7B Models}
To understand the model scaling possibilities of \model, in this section we compare \model 1.3B with its 2.7B variant, both trained on 1M steps. Table~\ref{tab:calame_1b_2b} shows the results, where it can be observed that the 2.7B is able to outperform the 1M steps 1.3B variant. 
This leads us to strongly believe that \model~performance has the potential to increase by scaling the model and by conducting further pre-training.

\section{Comparison to PT Encoder Models}
We now compare \model~with state-of-the-art PT encoder models on PT discriminative/non-generative tasks. 
In these tasks, classification/regression heads are added to the pre-trained model and fine-tuned in a fully supervised setting.
Previous research has shown that mostly due to their bidirectional nature, encoder models are particularly well-suited for many discriminative tasks, generally outperforming decoder-only models. For example, the GLUE leaderboard\footnote{\href{https://gluebenchmark.com/leaderboard}{GLUE Benchmark leaderboard}} is dominated by BERT-based models. 
In this section we compare  \model~to other PT-encoder models. While we know priori that this is not the setting in which decoders excel, it will allow us to understand how \model~positions itself against encoder approaches.

\subsection{Methodology Overview}
In the following evaluations, we considered the 1.3B version of \model~ and evaluated its 1M, 1.5M, 2M, and 3M step checkpoints.

For each task/subtask, we defined sets of hyperparameters to be evaluated (comprising learning rate, number of epochs, scheduler, etc.). Each model (including baselines) was fine-tuned in all hyperparameter sets,
using the same protocol. 
In tasks with multiple target metrics, for each experiment, we kept the best checkpoint for each metric, based on the validation set. We then report the results obtained with the best set of hyperparameters.
Furthermore, to increase robustness, each metric result was obtained by averaging the individual checkpoints' metric results. 

\subsection{ASSIN2}
ASSIN-2~\cite{assin2} is a PT-BR multitask benchmark whose goal is to train and evaluate models for assessing both entailment (RTE) and similarity (STS) relations between sentences. Its training, validation, and test sets comprise 6.5k, 500, and 3k sentence pairs with annotations for both tasks, respectively. Due to ASSIN-2 being PT-BR, we compared \model~ to BERTimbau-Large.

\begin{table}[t]
\centering
\caption{Best results achieved for each baseline, on the ASSIN-2 task, across all experiments. }
\resizebox{0.5\textwidth}{!}{%
\begin{tabular}{llll}
\toprule
\textbf{Model} & \textbf{F1} & \textbf{Accuracy} & \textbf{Pearson} \\ 
\midrule
\textbf{\model~1.3B} & 0.8960 & 0.8967 & \textbf{0.8510} \\
\textbf{BERTimbau-Large} & \textbf{0.9020} & \textbf{0.9020} & 0.8460 
\\ \bottomrule
\end{tabular}%
}
\label{tab:assin2topresults}
\end{table}

\begin{table*}[t]
\caption{
Evaluation results on the GLUE-PTPT tasks across all experiments (all fine-tunes). \textit{Enc.} stands for Encoders, and \textit{Dec.} stands for Decoders.}
\centering
\begin{tabular}{lccccc}
\toprule
 & \textbf{Models} & \textbf{RTE} & \multicolumn{2}{c}{\textbf{MRPC}} & \textbf{STS-B} \\ 
 &  & \textbf{Acc} & \textbf{F1} & \textbf{Acc} & \textbf{Pearson} \\
\midrule
\multirow{2}{*}{\begin{turn}{90}\textbf{Dec.}\end{turn}}   
& \model  & 0.6679 & 0.8775 & 0.8162 & 0.8500     \\
& Gervásio-PTPT  & 0.6534 & 0.8599 & 0.7941 & 0.8360 \\ \midrule
\multirow{2}{*}{\begin{turn}{90}\textbf{Enc.}\end{turn}}  
& Albertina-PTPT & \textbf{0.8628} & \textbf{0.9261} & \textbf{0.8971} & \textbf{0.898}    \\
& BERTimbau-Large & 0.6968 & 0.9030 & 0.8652 & 0.8700    \\ \bottomrule
\end{tabular}
\label{tab:glueptpttopresults}
\end{table*}

\vspace{2mm}\noindent\textbf{ASSIN-2 Protocols.\quad}
For the ASSIN-2 benchmark, we follow~\cite{BERTIMBAU} and perform a multi-task fine-tuning, by attaching two extra heads, each taking as input the embedding of the last token of the sequence. The final loss is the sum of the two losses from each task. RTE is treated as a classification task, thus we adopt the cross-entropy loss. STS is treated as a regression task, thus, we adopt the mean-squared error loss. To prepare the input, we tokenize the pair of sentences and pass the corresponding RTE and STS labels to the model, with a max sequence length of 128. \par
For this task's experimental space, we evaluated learning rates \textit{1e-5} and \textit{1e-6}, for 5 to 10 epochs, and for both linear and constant schedulers. A batch size of 32 was used with 2 GA steps. From these variations, we prepared 8 hyperparameter sets, and found that the most optimal combination for both our model and BERTimbau used a LR of \textit{1e-5}, 10 epochs, and a constant scheduler.

\vspace{2mm}\noindent\textbf{ASSIN-2 Results.\quad}
Table \ref{tab:assin2topresults} shows the best results from each model on the ASSIN-2 task.
A key observation is that \model~ achieves equivalent results to the encoder-based baseline, BERTimbau-large. In fact, our model achieves top-performance in terms of Pearson score, and comes very close to BERTimbau's F1 and Accuracy scores.

\subsection{Glue-PTPT} 
Given our focus on PT-PT, we evaluate \model~on GLUE-PTPT~\cite{albertinapt}, a PT-PT machine-translated version of GLUE~\cite{GLUE}. GLUE-PTPT comprises 4 subtasks of the original GLUE benchmark, from which we chose: RTE, MRPC, and STS-B. 
We compare \model~against Albertina-PTPT (encoder)~\cite{albertinapt} and Gervásio-PTPT (decoder)\footnote{\url{https://huggingface.co/PORTULAN} - model name: \texttt{gervasio-ptpt-base}.}.

\vspace{5mm}\noindent\textbf{Glue-PTPT Protocols.\quad}
Following the methodology, 4 hyperparameter sets were prepared for each subtask. The RTE and MRPC tasks share the same 4 sets - varying LR (\textit{1e-4} and \textit{1e-5}), linear and constant schedulers - while STS-B uses different ones - adding \textit{1e-6} as an extra LR value. For all subtasks, models were fine-tuned for 5 epochs, with a batch size of 32, and 2 gradient accumulation steps. For the input, each pair of sentences is tokenized with their corresponding label, with a max sequence length of 128, due to the sentences being relatively short. \par
At the time of writing, GLUE's official evaluation service was not available, so we followed Albertina's protocol~\cite{albertinapt} and used the original validation set as a test set, and took 10\% from the original train split to create a new validation split. 
All models and baselines were fine-tuned using the created splits, to ensure comparability.

\vspace{5mm}\noindent\textbf{Glue-PTPT Results.\quad}
The results, presented in Table \ref{tab:glueptpttopresults}, show that encoder-base models achieve better performance than decoder-based ones, with Albertina-PTPT achieving top performance followed by BERTimbau-large.
Nevertheless, among decoder-base models, \model~ significantly outperforms Gervásio-PTPT. This entails that among PT-PT decoder models, \model~ is a robust choice.

\section{Qualitative Results}
To complement quantitative evaluation, we conduct a qualitative evaluation of \model, by prompting the model to generate text for a set of topic-diverse prompts, using beam search w/ top-k sampling. The generated examples are illustrated in Table \ref{tab:textsamples}. 
The different generations showcase the model acquired knowledge across the different topics, ranging from \textit{Culinary}, \textit{Sports}, \textit{Health}, \textit{History}, etc. Namely, we observe that \model~\textbf{can output coherent and contextually correct PT-PT text}. In particular, the diversity of topics that we highlight in Table~\ref{tab:textsamples} hints that the model was able to capture the full range of topics that were present in the training data.

\begin{table*}[!ht]
\caption{\model~1.3B's text generation samples, with \textcolor{\promptcolor}{prompts} and \textcolor{\gencolor}{\textit{generations}}, using beam search w/ top-k (T=1.0) decoding. \textit{NL} represents a \textit{new line}.}
\centering
\small
\begin{tabularx}{\linewidth}{X|X}
\toprule
\textbf{Culinary:} \textcolor{\promptcolor}{A cozinha portuguesa é rica em aromas e sabores. Existem muitos pratos} \textcolor{\gencolor}{\textit{tradicionais, como o bacalhau com natas, o polvo à lagareiro ou a feijoada de búzios, mas também existem novos pratos que estão a ganhar cada vez mais adeptos. É o caso dos rojões à moda do}} (...)
&
\textbf{Tech-Auto:} \textcolor{\promptcolor}{Os carros são considerados uma das maiores criações da engenharia. Os avanços no transporte de seres humanos} \textcolor{\gencolor}{\textit{e máquinas, bem como as novas tecnologias, permitiram que o automóvel se tornasse um meio de transporte muito mais rápido e eficiente do que qualquer outro tipo de veículo.[NL] No entanto, a maior parte dos automóveis atuais ainda é baseada em motores de combustão interna}} (...)
\\ \midrule
\textbf{Sports:} \textcolor{\promptcolor}{ Surfing é um dos desportos mais praticados nas zonas costeiras de Portugal. Os adeptos do surf} \textcolor{\gencolor}{\textit{em Portugal são cada vez mais e o número de praticantes tem vindo a aumentar nos últimos anos.[NL] O surf é um desporto que se pratica com uma prancha, geralmente de madeira, para deslizar sobre as ondas. É praticado em todo o mundo,}} (...)
&
\textbf{Finance:} \textcolor{\promptcolor}{A literacia financeira permite às pessoas gerirem o seu dinheiro de forma mais responsável. Para aumentar a literacia financeira}\textcolor{\gencolor}{\textit{, é importante que as crianças e os jovens aprendam a gerir o seu dinheiro de forma adequada.[NL] Aprender a gerir o seu dinheiro vai ajudá-lo a:[NL] Compreender como funciona o dinheiro;[NL] Identificar diferentes fontes de rendimento}} (...)
\\ \midrule
\textbf{Environment:} \textcolor{\promptcolor}{Reciclar é muito importante para ajudar o ambiente. Reciclar} \textcolor{\gencolor}{\textit{garrafas de plástico, latas e frascos de vidro pode ser uma boa forma de ajudar o meio ambiente.[NL] Porquê reciclar?[NL] A reciclagem é um processo que tem como objetivo reaproveitar os materiais em fim de vida, transformando-os}} (...) & \textbf{History:} \textcolor{\promptcolor}{Antes do início do milénio, Portugal foi marcado por vários eventos marcantes, como por exemplo, em 1998}\textcolor{\gencolor}{\textit{, a Expo'98, que se realizou no Parque das Nações, em Lisboa. Este evento serviu para mostrar ao mundo o que de melhor se fazia na área da engenharia e construção civil em Portugal.[NL] A Expo'98 é um dos maiores eventos}} (...)
\\ \midrule
\textbf{Music:} \textcolor{\promptcolor}{A nível artístico, o legado musical português é rico. Alguns do maiores artistas portugueses são} \textcolor{\gencolor}{\textit{reconhecidos internacionalmente, como Amália Rodrigues, Carlos Paredes, José Afonso, Adriano Correia de Oliveira, José Mário Branco, Fausto, Sérgio Godinho, entre muitos outros. A nível internacional, destacam-se nomes como a Orquestra Sinfónica Portuguesa, a Orquestra Gulbenkian, a}} (...) &  \textbf{Health:} \textcolor{\promptcolor}{É importante proteger a nossa saúde. Um dos cuidados a ter é a toma de vacinas anuais. As vacinas} \textcolor{\gencolor}{\textit{são essenciais para a prevenção de doenças e, por isso, devemos tomá-las todos os anos.[NL] A vacina contra o tétano é uma das vacinas mais importantes para a proteção da nossa saúde. O tétano é uma doença que}} (...)
\\ \bottomrule
\end{tabularx}%
\label{tab:textsamples}
\end{table*}

\section{Discussion and Conclusions}
\subsection{Generative and Open Portuguese LLM}
In this paper we proposed \model, a generative and open large language model for Portuguese. In addition, we assemble a large-scale corpora for European Portuguese and contribute with CALAME-PT, a new benchmark for Portuguese generation tasks. 

\model~achieves state-of-the-art results in Portuguese generative tasks and is a competitive model on many discriminative tasks. 
We believe that this success is attributed to its larger size, training duration, and especially to its large and rich 35+ billion tokens corpora, comprising multiple high-quality PT-PT sources.

\subsection{Foundational Portuguese LLM and Broader Impact}
\model~establishes a strong foundation to pursue new advances in language modeling for European Portuguese.
Results demonstrated that \model~generates syntactically correct Portuguese language for a wide range of domains (Table~\ref{tab:textsamples}). 
The generated language is also semantically correct, with sentence structures demonstrating a sound knowledge about multiple topics with limited hallucinations. 
Despite lacking a structured knowledge training task, the model was able to produce semantically coherent generations, by inferring entities, their relations, and context.
For these reasons, we believe that \model~ model lays out a strong foundation to tackle complex NLP tasks requiring chain-of-thought, zero/few-shot reasoning, human alignment, among other challenging scenarios. 

\subsection{Limitations}
Our contributed CALAME-PT enables the evaluation of one particular facet of language modeling. However, the flexibility of such LLMs goes far beyond text completion, being capable of addressing tasks like abstractive summarization and dialog, either in zero or few-shot settings. Albeit such benchmarks are lacking for the Portuguese language, performing such evaluations would strengthen PT LLMs' research.

While the \model~ generated text is syntactically, grammatically, and contextually correct, similarly to LLMs in other languages, \textit{artifacts} may still be generated, including wrongly contextualized and non-factual generations. While some of these issues can be overcome with improved data selection~\cite{hallucination_survey}, carefully designed prompts~\cite{jin-etal-2022-good}, or constrained decoding strategies~\cite{rashkin-etal-2021-increasing}, further research is still required to mitigate this behavior, as these are challenges that go beyond PT LLMs.
Finally, while \model~is focused on European Portuguese, the ideal Portuguese LLM would cover other Portuguese variants as well (e.g. Mozambique, Guinea-Bissau, and others).
Such promising research directions are left for future work.

\subsection{Open Challenges}
The framework proposed in this paper enables tackling open LLM challenges.
This includes scaling the model to a larger number of parameters, including more training corpus, and expanding the model towards a multimodal LLM~\cite{liu2023llava}.
In addition, \model~enables bringing new learning paradigms to Portuguese language modeling, such as LLM human-aligned generation: instruction tuning~\cite{rlhf,dpo}, factuality~\cite{llm_factuality}, and dialog~\cite{plangpt,twiz_2023}.

\section*{Acknowledgements}
We would like to thank Arquivo.pt's team for their content preservation efforts, and for all the help and guidance in accessing the archived web pages at scale.
This work has been partially funded by the FCT project NOVA LINCS Ref. UIDP/04516/2020, by CMU|Portugal project iFetch, Ref. CMUP LISBOA-01-0247-FEDER-045920, and by the FCT project Ref. Nº CPCA-IAC/AV/594875/2023.

\clearpage
\bibliography{custom}

\appendix

\end{document}